# Algorithmic Analysis of Edge Ranking and Profiling for MTF Determination of an Imaging System

Poorna Banerjee Dasgupta

*M.Tech Computer Science & Engineering, Nirma Institute of Technology*
*Ahmedabad, Gujarat, India*

*Abstract*— **Edge detection is one of the most principal techniques for detecting discontinuities in the gray levels of image pixels. The Modulation Transfer Function (MTF) is one of the main criteria for assessing imaging quality and is a parameter frequently used for measuring the sharpness of an imaging system. In order to determine the MTF, it is essential to determine the best edge from the target image so that an edge profile can be developed and then the line spread function and hence the MTF, can be computed accordingly. For regular image sizes, the human visual system is adept enough to identify suitable edges from the image. But considering huge image data-sets, such as those obtained from satellites, the image size may range in few gigabytes and in such a case, manual inspection of images for determination of the best suitable edge is not plausible and hence, edge profiling tasks have to be automated. This paper presents a novel, yet simple, algorithm for edge ranking and detection from image data-sets for MTF computation, which is ideal for automation on vectorised graphical processing units.**

## I. INTRODUCTION

Edges are discontinuities in the gray levels of image pixels. An edge is a set of connected pixels that lie on the boundary between two regions. Blurred edges tend to be thick and sharp edges tend to be thin. The intensity profiles for an ideal edge and a step 'ramp' edge are shown in Figure 1[1],[7].

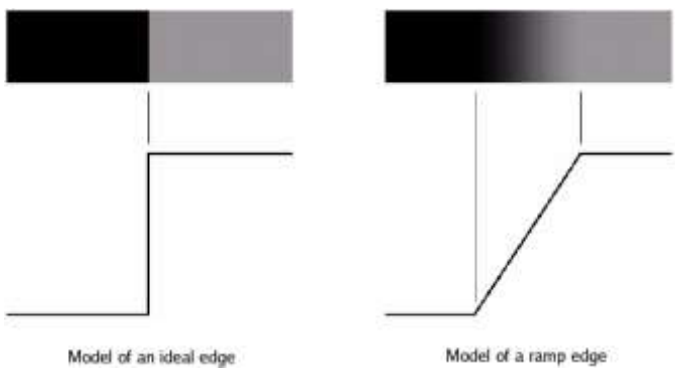

Fig. 1 Edge profiles for an ideal and ramp edge

The task of edge detection can be carried out by convolving gradient-based filters with the input image in both spatial and frequency domains. Some simple first-derivative based filters include Sobel, Prewitt and Robert's filters while Laplacian-of-Gaussian and Canny-edge detection are more advanced edge detection techniques.

Modulation Transfer Function (MTF) is a crucial criterion for assessing the quality and sharpness of an imaging system [2],[3]. Measuring the MTF is essential to carry out the focusing of a telescope, or to implement a deconvolution filter whose goal is to enhance the image contrast or reduce the noise. Its knowledge also helps us to compare the characteristics of different known and unknown imaging systems. There are several methods for determining the MTF of an imaging system. MTF is mathematically given as the Discrete Fourier Transform (DFT) of the Line Spread Function (LSF). The LSF can be calculated as the first derivative of the Edge Spread Function (ESF). In order to develop the ESF, it is essential to determine the best edge from the target image so that an edge profile can be developed and then the line spread function and hence the MTF, can be determined.

As mentioned earlier in the abstract of this paper, for small sized images, manual inspection is sufficient to determine suitable edges for profiling but in case of huge image data-sets (such as those involved with satellite imaging), the process for edge profiling must be automated. Such edge profiling tasks are ideal for automation on Graphics Processing Units (GPU).

However, the process of automated edge profiling for huge image data-sets brings into focus certain issues such as:

- The edge detection algorithm that should be used depends on factors such as the image thresholding levels, acceptable noise level in the image, the target imaging system, etc.
- Once all the edges have been detected in the target image, *some algorithm has to be devised for ranking the detected edges, so that the best edge can be chosen*. Some such parameters for ranking edges, *specifically for MTF determination* include: **thickness of the edges** (the ideal edge is one pixel thick), **length of the edges** (typical edges have the length of 100 or more pixels), and **orientation of the edges** for MTF determination (vertical or near-vertical edges are preferred).
- After determining the best ranked edge, the location of such an edge in the image also has to be determined, so that the best edge, along with its location and its edge profile can be provided as the algorithm's output. The

- edge-ranks can be stored in a vector array. Similar arrays or matrices can be constructed for storing other edge profile information and can be supplied with the output.
- After the rank vector has been constructed, it is necessary to apply efficient sorting algorithms in order to find the maximum rank, and hence the best edge, from the rank vector array.

Considering all the issues described above, this paper presents an implementable algorithm for automating edge profiling and edge-rank determination. The subsequent sections of this paper elucidate the proposed algorithmic formula for edge profiling & edge-rank determination and show the results of implementation of the algorithm on different sample image data-sets.

## II. EDGE-RANK DETERMINATION & EDGE PROFILING

In the proposed algorithm for edge-rank determination during MTF computation, three prime parameters have been considered for determining the rank (***R***) of an edge: the edge thickness (***t***), edge length (***l***) and the edge angle made with the x-axis (***Θ***). Using these three parameters, the following formula has been proposed for determining the rank of an edge:

$$R = (l-t)/10 + abs(\Theta)$$

where l, t are in units of pixels, $\Theta$ is in radians and $0 \leq \Theta \leq \pi/2$.

From the above formula it can be seen that suitable edges for edge profiling during MTF computation should be sufficiently long (100 pixels or more) and should ideally be single pixel thick and vertical or near-vertical edges are preferred. The edge angle $\Theta$ can be calculated with the help of the image gradient [1]. The gradient of an image $f(x,y)$ at location $(x,y)$ in the image is given by the vector $\nabla f$.

$$\nabla f = \begin{bmatrix} G_x \\ G_y \end{bmatrix} = \begin{bmatrix} \frac{\partial f}{\partial x} \\ \frac{\partial f}{\partial y} \end{bmatrix}$$

$$\nabla f = mag(\nabla f) = \left[G_x^2 + G_y^2\right]^{1/2}$$

If ***α(x,y)*** represents the direction angle of the image vector $\nabla f$ at point *(x,y)* of the image, then:

$$\alpha(x,y) = tan^{-1}(G_y/G_x)$$

The direction of an edge at *(x,y)* is perpendicular to the direction of the gradient vector at that point.

In order to carry out the task of edge profiling, first all edges from the test image have to be detected. For the purpose of demonstration in this paper, a 5x5 Laplacian-of-Gaussian filter was convolved with the test images in spatial domain for edge detection. Once the edges have been detected, binary thresholding can be applied in order to eliminate weak or false detected edges such as those resulting from noise. Keeping in mind the requirements for MTF computation, preferred edges would be nearly vertical or vertical. Applying binary thresholding makes it easy to finding out the starting coordinates and length of such edges during pixel scanning of the image. After this selective scanning of edges, the rank of each edge can be determined using the proposed formula.

The proposed formula was used for edge ranking and then edge profiling was carried out for sample images. Two such sample test images, along with the corresponding edge ranks, lengths and starting coordinates of the edges found are shown in Figures 2 and 3. The first sample image depicts a typical satellite image of a city landscape. Edge detection on such images help in detecting major roads, rivers and other important geographical features. The second sample image was chosen keeping in mind that vertical edges are preferred for MTF determination. The best edge found, i.e. the edge with the highest rank has been highlighted in a blue circle.

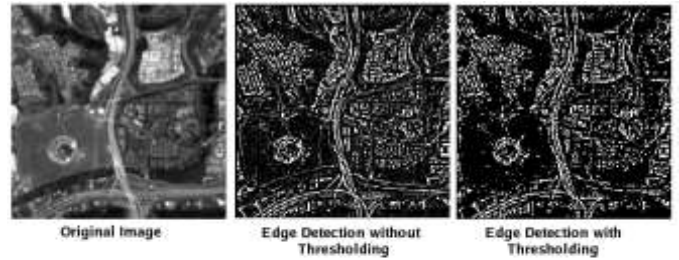

| Edge Rank | Edge Length | Starting Coordinates |
|---|---|---|
| 1.9961 | 6 | (123,6) |
| 0.9976 | 6 | (124,6) |
| 1.6956 | 7 | (117,18) |
| 1.8036 | 6 | (116,45) |
| 2.9109 | 15 | (115,49) |
| 2.5943 | 14 | (116,49) |
| 1.9172 | 13 | (117,49) |
| 1.3041 | 12 | (118,49) |
| 2.0296 | 11 | (119,49) |
| 2.1978 | 10 | (118,59) |
| 1.8505 | 9 | (122,74) |
| 1.1059 | 8 | (123,79) |
| …. | …. | …. |

Fig. 2 Edge Detection and Ranking for sample-image1

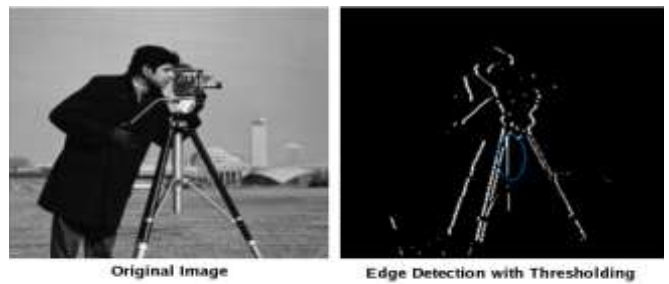

| Edge Rank | Edge Length | Starting Coordinates |
|---|---|---|
| 1.1218 | 8 | (168,72) |
| 1.1636 | 7 | (169,72) |
| 0.9218 | 6 | (170,72) |
| 0.8974 | 7 | (170,77) |
| 0.9218 | 6 | (172,80) |
| 1.1636 | 7 | (173,80) |
| 0.7974 | 6 | (163,80) |
| 0.7384 | 7 | (130,131) |
| 0.6768 | 6 | (131,131) |
| 1.6156 | 16 | (130,132) |
| 1.5312 | 15 | (131,132) |
| 1.4121 | 14 | (132,132) |
| …. | …. | …. |

Fig. 3 Edge Detection and Ranking for sample-image2

From Figures 2 and 3, it can be seen that the best found edge in the first image has a rank of 2.91 and edge length of 15

pixels and its starting coordinates is (115,49); in the second image the best found edge has a rank of 1.6156 with an edge length of 16 pixels and starting coordinates at (130,132). It might be noted that although the proposed edge-rank formula was designed keeping in mind huge image data-sets obtained through satellite imagery, the proposed algorithm and formula work just as well for small-sized images too.

In order to perform the same algorithm for large image data-sets, an iterative approach would be appropriate where the algorithm divides the input data-set into sub-images that can be fitted into the GPU memory and processes each sub-image per iteration with the help of multiple parallelized execution streams so that input-output (I/O) memory transfers between the host processor and the GPU can be overlapped asynchronously with kernel code execution. Execution of the proposed algorithm on general purpose GPU's (especially those GPU's with compute capability of 2.0 or higher) helps in tremendous speed-up of the execution time in comparison to normal sequential execution on multi-core host processors[4],[5],[6].

### III. CONCLUSIONS & FUTURE SCOPE OF WORK

Edge profiling is an indispensable aspect of Modulation Transfer Function (MTF) computation for an imaging system. This paper proposed and explored how the process of edge profiling can be automated for large image data-sets, keeping in mind the requirements for MTF computation, where manual inspection for finding the best suited edge from an image is not plausible. The results of implementation of the proposed algorithm were then demonstrated for two test images.

As further extensions to the research work carried out in this paper, sub-pixel accuracy can be taken into account while detecting edges. The edge detection and ranking algorithm can also be enhanced to convolve edge-detection filters in the frequency domain instead of the spatial domain.

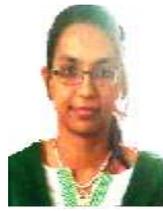

AUTHOR'S PROFILE:
**Poorna Banerjee Dasgupta** has received her B.Tech & M.Tech Degrees in Computer Science and Engineering from Nirma Institute of Technology, Ahmedabad, India. She did her M.Tech dissertation at Space Applications Center, ISRO, Ahmedabad, India and has also worked as Assistant Professor in Computer Engineering dept. at Gandhinagar Institute of Technology, Gandhinagar, India from 2013-2014 and has published research papers in reputed international journals. Her research interests include image processing, high performance computing, parallel processing and wireless sensor networks